\documentclass[10pt, a4paper]{article}

\usepackage{lrec2026} 

\usepackage{booktabs} 

\usepackage{amsmath}

\usepackage{xurl}

\usepackage{caption}

\usepackage{graphicx} 

\usepackage{colortbl}

\usepackage{xcolor}
\definecolor{baseline}{HTML}{D6E4F0}
\definecolor{sweetlora}{HTML}{C6EFCE}

\newcommand{\sweetval}[1]{\cellcolor{sweetlora}\textbf{#1}}

\definecolor{errorred}{rgb}{0.75, 0.0, 0.0}
\definecolor{fixgreen}{rgb}{0.0, 0.50, 0.0}

\definecolor{sweetspot}{gray}{0.90}

\usepackage{multirow}

\title{On the Role of Encoder Depth: Pruning Whisper and LoRA Fine-Tuning in SLAM-ASR}

\name{Ganesh Pavan Kartikeya Bharadwaj Kolluri, Michael Kampouridis, Ravi Shekhar} 

\address{School of Computer Science and Electronic Engineering, University of Essex \\  
\texttt{\{karthik.kolluri, mkampo, r.shekhar\}@essex.ac.uk}}

\abstract{Automatic speech recognition (ASR) has advanced rapidly in recent years, driven by large-scale pretrained models and end-to-end architectures such as SLAM-ASR. A key component of SLAM-ASR systems is the Whisper speech encoder, which provides robust acoustic representations. While model pruning has been explored for the full Whisper encoder–decoder architecture, its impact within the SLAM-ASR setting remains under-investigated. In this work, we analyze the effects of layer pruning in the Whisper encoder when used as the acoustic backbone of SLAM-ASR. We further examine the extent to which LoRA-based fine-tuning can recover performance degradation caused by pruning. Experiments conducted across three Whisper variants (Small, Medium, Large-v2), three languages representing distinct resource levels (Danish, Dutch, English), and over 200 training runs demonstrate that pruning two encoder layers causes only 2–4\% WER degradation, and that combining this pruning with LoRA adaptation consistently outperforms the unpruned baseline while reducing total parameters by 7–14\%. Moreover, our error analysis reveals that LoRA primarily compensates through the language model's linguistic priors, reducing total word errors by 11-21\% for Dutch and English, with substitutions and deletions showing the largest reductions. However, for low-resource Danish, the reduction is smaller (4-7\%), and LoRA introduces increased insertion errors, indicating that compensation effectiveness depends on the LLM's pre-existing language proficiency and available training data.
\\ \newline 
}

\begin{document}

\maketitleabstract

\section{Introduction}
\label{sec:intro}

Recent advances in multimodal models have enabled automatic speech recognition systems (ASR) through architectures that connect pre-trained speech encoders to large language models (LLMs). SLAM-ASR \citep{ma2024embarrassingly, zhang-etal-2026-language, zhang-etal-2026-speak} is the one such instance, which is a simpler approach and yet achieved comparative ASR performance: a frozen speech encoder extracts acoustic representations, and a small trainable projector maps these into the embedding space of a frozen LLM, which then generates transcriptions autoregressively. While this modular design is attractive for its simplicity, deploying such systems remains computationally expensive, as the full encoder must process every input audio regardless of whether all its layers contribute meaningfully to ASR performance. 

Model compression through layer pruning has been widely studied for ASR. \citet{kim2023adapt} combined language-specific adapters with magnitude pruning to compress Whisper by up to 50\%, still with competitive CER performance, while \citet{gu2024sparsewav} achieved 80\% compression through one-shot unstructured pruning with minimal WER degradation. On the decoder side, \citet{gandhi2023distil} distilled Whisper's decoder from 32 layers to 2 using knowledge distillation with large-scale pseudo-labeling, achieving 6 times faster inference within 1\% WER of the original. However, these studies focused on traditional encoder-decoder ASR, where both the encoder and decoder are jointly trained. In the SLAM-ASR case, the encoder remains the same; the decoder is replaced by a frozen LLM that was never exposed to speech data and a lightweight projector that bridges the two components together. The effects of encoder layer pruning on this architecture remain unexplored. Layer-wise analyses suggest that upper encoder layers encode increasingly abstract linguistic features \citep{pasad2021layer}, capabilities the LLM already possesses, raising the possibility that these layers are partially redundant. Separately, Low-Rank Adaptation (LoRA) \citep{hu2022lora} has been applied to LLM-based ASR for multilingual adaptation \cite{song2024lora, fang2025low, concina25_mlcslm} and low-resource improvement \cite{ma2024embarrassingly,nagano2025llm, fong2025speech, Burdisso2026text}, but no prior work has examined whether LoRA can compensate for performance lost through structured encoder layer pruning, or how data resource availability regulates these interactions. 

This paper addresses these gaps through a systematic study of encoder layer pruning and LoRA compensation in SLAM-ASR \footnote{The code is available at \url{https://github.com/KarthikKolluriKB/SLAM-ASR-Encoder-Pruning-LoRA}
}. Specifically, we focus on two research questions: 

RQ1: \textit{How does progressive encoder layer pruning affect ASR performance, and how does data resource availability modulate this effect?}

RQ2: \textit{Can LoRA adaptation compensate for pruning induced degradation?}

To answer these questions, we evaluated across three languages representing distinct resource levels: Danish (4.2 hours, low-resource), Dutch (50 hours, medium-resource), and English (100 hours, high-resource), using three Whisper encoder variants (Small, Medium, Large-v2) \citep{radford2023robust} paired with Qwen2.5-3B LLM \citep{qwen2025qwen25technicalreport}. All three languages are supported by both the Whisper encoders and Qwen2.5-3B, ensuring that observed differences reflect data availability rather than missing language coverage. Our experiments cover pruning depths from the full encoder down to a single layer, with and without LoRA, spanning over 200 individual training runs.

Our main findings are as follows: First, removing the top one to two encoder layers results in only marginal WER degradation (with 2-4\%) across all encoder scales and languages, supporting the hypothesis that upper encoder layers are partially redundant when an LLM handles linguistic processing. Beyond this arrangement, medium and high-resource languages degrade smoothly, while low-resource Danish exhibits erratic, non-monotonic behaviour. Second, combining modest pruning (two layers) with LoRA consistently outperforms the unpruned baseline while reducing total parameters by 7–14\%, demonstrating that fewer encoder parameters paired with lightweight LLM adaptation can match or exceed full baseline model performance. Third, LoRA's compensatory benefit is not uniform across resource levels: it provides robust improvement for Dutch and English across all pruning depths, but is less effective for Danish, suggesting an interaction between data availability and adaptation capacity that needs to be further investigated. Additionally, error analysis reveals that LoRA compensates primarily through the LLM's linguistic priors, reducing word-level errors by 11–21\% for Dutch and English, with substitution and deletion corrections showing the largest reductions, confirming a decoding-side compensation mechanism.

\section{Related Works}
\label{sec:relWork}

This section reviews prior research in two areas relevant to our study: (1) layer pruning strategies for speech encoders, and (2) parameter-efficient fine-tuning for LLM-based speech systems.

\subsection{Layer Pruning in Speech Encoders}

Model compression through layer pruning has been extensively studied for ASR. \citet{kim2023adapt} introduced PEPSI (Parameter-Efficient Pruning and Adaptation for Speech Foundational Models), an adapt-and-prune framework for Whisper that combines language-specific adapters with iterative magnitude pruning. Their experiments on Common Voice showed that pruning can reduce model size by up to 50\% while maintaining competitive CER across Korean and Malayalam. However, PEPSI employs \textit{unstructured} pruning that removes individual weights rather than entire layers, resulting in sparse networks that require specialized hardware for efficient inference.

More recently, \citet{gu2024sparsewav} proposed SparseWAV, a one-shot unstructured pruning method for large speech foundation models, achieving up to 80\% compression with minimal WER degradation. \citet{irigoyen2025pruning} revealed that certain encoder components actually \textit{improve} when pruned, acting as implicit regularizers. Decoder self-attention at 50\% sparsity achieved 2.38\% absolute WER reduction on LibriSpeech test-other.

A parallel line of study focuses on \textit{decoder} compression. Distil-Whisper \citep{gandhi2023distil} uses knowledge distillation \citep{hinton2015distilling} with large-scale pseudo-labeling to compress the decoder from 32 layers to 2 while keeping the encoder frozen, achieving 6$\times$ faster inference within 1\% WER of the original model. Similarly, BaldWhisper \citep{sy2025baldwhisper} targets low-resource deployment by merging decoder layer pairs in Whisper-base rather than removing them, achieving 2.15$\times$ faster inference with 48\% size reduction while maintaining over 90\% of baseline performance on Bambara speech data.

However, these studies focus on traditional encoder-decoder ASR, where Whisper's own decoder directly generates transcriptions. SLAM-ASR \citep{ma2024embarrassingly} presents a fundamentally different setting: the decoder is replaced by an LLM, and a lightweight projector that maps encoder outputs to the LLM embedding space. The effects of encoder layer pruning on this SLAM-ASR architecture remain unexplored. Here, the projector must learn to align potentially degraded encoder representations with a fixed, independently pre-trained LLM. Unlike traditional ASR, where encoder and decoder are jointly trained, SLAM-ASR offers no such flexibility: the LLM is frozen and was never exposed to speech, placing the entire burden of representation quality on the encoder. 

\subsection{Parameter-Efficient Fine-Tuning for LLM-based Speech Systems}

Parameter-efficient fine-tuning (PEFT) methods \citep{houlsby2019parameterefficienttransferlearningnlp} have become essential for adapting pre-trained large language models to specific downstream tasks without updating the entire model’s parameters. Low-Rank Adaptation (LoRA) \citep{hu2022lora} has emerged as the dominant approach, decomposing weight updates into low-rank matrices that typically account for less than 1\% of total parameters.

In LLM-based ASR, LoRA has been applied across multiple components. \citet{song2024lora} proposed LoRA-Whisper, which incorporates the LoRA matrix into Whisper for multilingual ASR to effectively mitigate language interference, achieving 18.5\% relative WER reduction for multilingual ASR and 23.0\% for language expansion while using only 5\% of trainable parameters compared to full fine-tuning. Building on this idea, \citet{mu2025hdmole} proposed HDMoLE, which combines multiple LoRA experts through hierarchical routing and dynamic thresholds for multi-accent LLM-based ASR. Their method achieves comparable CER to full fine-tuning on multi-accent and standard Mandarin datasets while using only 9.6\% of the trainable parameters.

Within LLM-based ASR models, the SLAM-ASR framework \citep{ma2024embarrassingly} showed that applying LoRA to LLM attention layers improves low-resource ASR, especially when combined with partial encoder fine-tuning \citep{tang2023salmonn, wu2023decoder}. However, \citet{kumar2025performance} observed that SLAM-ASR generalizes poorly across domains, highlighting the need for effective adaptation strategies in real-world deployment. 

Despite this progress, no prior work has examined whether LoRA can compensate for information lost through structured encoder layer removal. While PEPSI \citep{kim2023adapt} combines LoRA with unstructured weight pruning, the effects of removing entire encoder layers and whether LoRA can recover from such architectural changes in SLAM-ASR systems remain unexplored. It is also unknown whether LoRA's compensatory effectiveness depends on the amount of available training data.

\section{Methodology}
\label{sec:method}

This study investigates two strategies for improving the parameter efficiency of SLAM-ASR systems: encoder layer pruning and LoRA adaptation of the LLM. Specifically, we ask whether upper encoder layers can be removed without significant performance loss, given that the LLM is already proficient in linguistic tasks, and whether LoRA can compensate for any degradation introduced by pruning. We evaluated these strategies across three languages representing distinct resource levels to examine how data availability interacts with both pruning robustness and LoRA effectiveness. 

\begin{figure}[t]
\centering
\includegraphics[width=0.90\columnwidth]{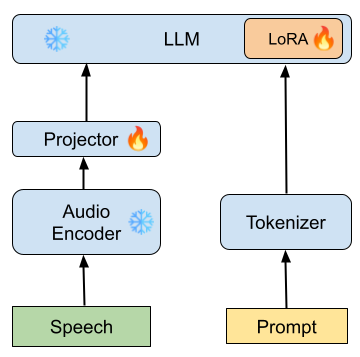}
\caption{Overview of the SLAM-ASR architecture.}
\label{fig:slam_asr}
\end{figure}

We follow the SLAM-ASR framework \citep{ma2024embarrassingly}, which connects a frozen pre-trained speech encoder to a frozen large language model through a small trainable projector module. Only projector parameters are updated during training, and both the encoder and LLM remain frozen. This modular design is crucial for our study because each component has a definite role; we can selectively modify the encoder through pruning and study how the system responds without disrupting other components. Figure~\ref{fig:slam_asr} illustrates this architecture.

For encoder pruning, these previous studies on layer-wise analysis have shown that lower encoder layers primarily capture low-level acoustic features, while upper layers encode increasingly abstract linguistic information \citep{pasad2021layer}. We hypothesize that in the SLAM-ASR architecture, these upper layers may be partially redundant, and that removing them offers a favourable efficiency–performance trade-off by delegating linguistic processing to the LLM. To verify this, we adopt a top-down pruning strategy: starting from the full encoder, we sequentially remove layers from the top, producing increasingly compressed configurations. For each configuration, we retrain only the projector from scratch, keeping both the pruned encoder and LLM frozen. 

To investigate whether the LLM can be adapted to better handle degraded encoder representations, we apply Low-Rank Adaptation (LoRA) \citep{hu2022lora} to all attention projection matrices (Q, K, V, O) in the LLM. We hypothesize that, rather than recovering missing encoder information directly, LoRA enables the LLM to leverage its pre-trained linguistic knowledge: vocabulary, grammar, and contextual plausibility to adapt to the lost acoustic details. This creates a parameter trade-off: we remove millions of encoder parameters while adding far fewer LoRA parameters. We evaluate all combinations of pruning depths with and without LoRA, using Word Error Rate (WER) as the primary evaluation metric.

\section{Experimental Setup}
\label{sec:exp}

\subsection{Datasets}

We evaluate our approach on the Common Voice 22 corpus \citep{ardila2020common}, a crowdsourced multilingual speech dataset. To study how encoder pruning and LoRA compensation interact with training data availability, we select three languages that span an order-of-magnitude range in training data: Danish (4.2 hours), Dutch (54 hours), and English (100 hours). This selection is intentional; these three resource levels allow us to test whether pruning degradation patterns are consistent across data resource availability (RQ1) and whether LoRA compensation depends on sufficient fine-tuning data (RQ2). Dataset statistics are summarised in Table~\ref{tab:dataset_stats}.

\begin{table}[h]
\centering
\caption{Dataset statistics for each language from Common Voice 22.}
\label{tab:dataset_stats}
\begin{tabular}{llrr}
\toprule
\textbf{Language} & \textbf{Split} & \textbf{Samples} & \textbf{Hours} \\
\midrule
Danish  & Train & 3,592  & 4.18  \\
        & Dev   & 2,511  & 3.21  \\
        & Test  & 2,684  & 3.45  \\
\midrule
Dutch   & Train & 43,458 & 54.15 \\
        & Dev   & 12,032 & 16.03 \\
        & Test  & 12,033 & 16.38 \\
\midrule
English & Train & 58,140 & 100.00 \\
        & Dev   & 16,398 & 27.22  \\
        & Test  & 16,391 & 27.02  \\
\bottomrule
\end{tabular}
\end{table}

\paragraph{Preprocessing.} Audio files are resampled to 16\,kHz and converted to 80-channel log-Mel spectrograms following the Whisper preprocessing pipeline. We filter utterances to retain only those between 0.5 and 30 seconds in duration. Transcriptions are lowercased with punctuation removed, preserving apostrophes for contractions, following standard ASR preprocessing conventions.

\subsection{Implementation Details}

\paragraph{Model Architecture.} We use Whisper \citep{radford2023robust}, a transformer-based \citep{vaswani2017attention} speech encoder, as it is a widely adopted multilingual encoder available in multiple model sizes, making it well-suited for studying how pruning dynamics vary with encoder capacity. We pair it with Qwen2.5-3B \citep{qwen2025qwen25technicalreport} as the LLM backbone, selected for its multilingual pre-training coverage, which is essential given that our experiments span three distinct languages. We experiment with three Whisper variants to study how pruning dynamics vary with model capacity. For all three Whisper variants: Small (12 layers), Medium (24 layers), and Large-v2 (32 layers), we evaluate configurations down to 2 layers in increments of 2. 

\paragraph{Projector Architecture.} We employ a two-layer MLP with concatenation-based temporal downsampling. The projector concatenates 5 consecutive frames (reducing the sequence length by a factor of 5), then applies two linear transformations with ReLU activation and a dropout rate of 0.1. The hidden dimension is 2048. We apply LayerNorm after the final projection to match the scale of LLM text embeddings.

\paragraph{LoRA Configuration.} We apply Low-Rank Adaptation \citep{hu2022lora} to the query, key, value, and output projection matrices of all LLM attention layers. To prevent overfitting on smaller datasets, we use separate configurations for each resource level: $r{=}8$, $\alpha{=}16$ for Danish (low-resource) and $r{=}16$, $\alpha{=}32$ for Dutch/English (medium/high-resource). In preliminary experiments, using r=16 for Danish led to overfitting, motivating the reduced rank. All LoRA modules use a dropout of 0.1, adding approximately 0.8M (r=8) and 1.5M (r=16) trainable parameters, respectively.

\subsection{Training Details}

All models are trained using AdamW \citep{loshchilov2017decoupled} with learning rate $1{\times}10^{-4}$, weight decay 0.01, and gradient clipping at 1.0. We employ a cosine learning rate schedule with linear warmup over the first 5\% of training steps. Training uses mixed-precision (bfloat16) on a single NVIDIA A6000 GPU (48GB). Both the Whisper encoder and the Qwen2.5-3B base weights remain frozen; only the projector weights (and LoRA adapters when enabled) are updated.

\subsection{Evaluation Details}

We report Word Error Rate (WER) on the held-out test sets of each language. During evaluations, we use beam search with a beam size of 2. Text normalisation is applied to both predictions and references before evaluation: all text is lowercased, and punctuation is removed. For Danish, special characters (æ, ø, å) are preserved during normalisation to avoid penalising correct transcriptions. WER is computed using the \texttt{jiwer} library \footnote{\url{https://github.com/jitsi/jiwer}}. 

\section{ Results and Discussion}
\label{sec:results}

This section presents findings organized around two research questions: (1) How does progressive encoder layer pruning affect ASR performance, and how does data availability modulate this effect? (Section~\ref{sec:pruning_results}); and (2) Can LoRA adaptation compensate for pruning-induced degradation? (Section~\ref{sec:lora_results}). We further provide an error analysis examining per-utterance trade-offs and the mechanism underlying LoRA compensation (Section~\ref{sec:error_analysis}) 

\begin{table*}[t]
\centering
\setlength{\tabcolsep}{10pt}
\caption{Test set WER (\%) across encoder layer pruning configurations for three Whisper scales. 
\textit{Red.\%}: reduction of parameters in percentage. 
\textit{Base}: projector-only training; \textit{+LoRA}: LoRA additionally applied to the LLM. 
\colorbox{baseline}{Blue}: full baseline WER.
\colorbox{sweetlora}{\textbf{Green}}: 2 layer pruned+LoRA that outperforms the full baseline.}
\label{tab:layer_pruning_wer}
\begin{tabular}{@{} l c rr rr rr @{}}
\toprule
 & & \multicolumn{2}{c}{\textbf{Danish}} & \multicolumn{2}{c}{\textbf{Dutch}} & \multicolumn{2}{c}{\textbf{English}} \\
\cmidrule(lr){3-4} \cmidrule(lr){5-6} \cmidrule(lr){7-8}
Layers & Red.\% & Base & +LoRA & Base & +LoRA & Base & +LoRA \\
\midrule
\multicolumn{8}{c}{\textbf{Whisper-Small (88.15M base params)}} \\ \hline
12 (full) & 0\%    & \cellcolor{baseline}47.41  & 41.94 & \cellcolor{baseline}17.98 & 15.40 & \cellcolor{baseline}21.84 & 17.87 \\
10       & 16.1\% & 50.49  & \sweetval{47.23} & 21.16 & \sweetval{16.98} & 25.81 & \sweetval{20.64} \\
8        & 32.2\% & 71.22  & 51.20          & 27.25 & 21.04          & 33.15 & 24.40 \\
6        & 48.2\% & 115.68 & 67.02          & 32.14 & 24.45          & 40.81 & 31.85 \\
4        & 64.3\% & 82.56  & 72.60          & 45.62 & 31.69          & 74.32 & 41.73 \\
2        & 80.4\% & 123.21 & 115.48         & 62.28 & 48.97          & 110.85 & 95.30 \\
\midrule
\multicolumn{8}{c}{\textbf{Whisper-Medium (307.24M base params)}} \\\hline
24 (full) & 0\%    & \cellcolor{baseline}39.00  & 38.88 & \cellcolor{baseline}12.74 & 10.92 & \cellcolor{baseline}16.33 & 14.13 \\
22       & 8.2\%  & 39.73  & \sweetval{37.91} & 14.22 & \sweetval{11.84} & 17.89 & \sweetval{15.82} \\
20       & 16.4\% & 43.18  & 41.71          & 15.73 & 13.44          & 18.29 & 16.54 \\
18       & 24.6\% & 50.59  & 44.74          & 18.25 & 14.66          & 20.44 & 18.47 \\
16       & 32.8\% & 49.72  & 49.20          & 20.58 & 15.76          & 23.15 & 20.22 \\
14       & 41.0\% & 55.55  & 51.95          & 22.64 & 16.91          & 25.85 & 21.79 \\
12       & 49.2\% & 59.15  & 59.21          & 27.61 & 20.04          & 29.44 & 25.38 \\
10       & 57.4\% & 63.21  & 62.02          & 27.85 & 24.50          & 33.79 & 27.70 \\
8        & 65.6\% & 69.62  & 66.03          & 38.16 & 27.89          & 43.79 & 34.03 \\
6        & 73.8\% & 77.08  & 79.41          & 44.00 & 34.27          & 97.35 & 72.30 \\
4        & 82.0\% & 105.16 & 102.63         & 53.40 & 43.46          & 111.71 & 77.48 \\
2        & 90.2\% & 148.37 & 90.25          & 66.83 & 56.23          & 110.12 & 78.01 \\
\midrule
\multicolumn{8}{c}{\textbf{Whisper-Large-v2 (636.83M base params)}} \\ \hline
32 (full) & 0\%     & \cellcolor{baseline}36.09  & 34.07 & \cellcolor{baseline}12.05 & 10.32 & \cellcolor{baseline}13.36 & 11.58 \\
30       & 6.2\%  & 38.36  & \sweetval{35.81} & 13.61 & \sweetval{11.44} & 14.87 & \sweetval{12.16} \\
28       & 12.4\% & 40.08  & 39.12          & 15.75 & 12.67          & 16.26 & 15.62 \\
26       & 18.5\% & 48.28  & 44.75          & 17.15 & 14.03          & 18.32 & 17.08 \\
24       & 24.7\% & 52.02  & 47.63          & 18.96 & 15.23          & 20.75 & 20.13 \\
22       & 30.9\% & 52.44  & 56.24          & 21.31 & 16.71          & 22.31 & 21.45 \\
20       & 37.1\% & 130.99 & 59.06          & 26.00 & 18.73          & 36.34 & 32.92 \\
18       & 43.3\% & 73.87  & 61.64          & 28.92 & 22.06          & 48.38 & 39.86 \\
16       & 49.4\% & 67.48  & 64.40          & 38.82 & 24.87          & 54.12 & 46.28 \\
14       & 55.6\% & 77.97  & 73.77          & 45.21 & 29.81          & 62.45 & 59.82 \\
12       & 61.8\% & 115.51 & 84.08          & 59.81 & 39.03          & 74.42 & 65.36 \\
10       & 68.0\% & 104.79 & 87.67          & 59.60 & 43.42          & 84.06 & 74.17 \\
8        & 74.2\% & 160.15 & 90.33          & 63.61 & 48.18          & 88.33 & 77.34 \\
6        & 80.3\% & 120.43 & 99.58          & 90.60 & 51.14          & 98.28 & 87.84 \\
4        & 86.5\% & 118.30 & 122.81         & 72.41 & 62.04          & 104.34 & 92.24 \\
2        & 92.7\% & 123.52 & 132.03         & 83.75 & 74.39          & 108.47 & 96.20 \\
\bottomrule
\end{tabular}
\end{table*}
\vspace{-0.8em}

\begin{table*}[t]
\centering
\renewcommand{\arraystretch}{1.2}
\caption{Sweet-spot configurations (bold) per encoder scale, showing WER (\%) on the test set. Net $\Delta$ is relative to the full baseline. Reported params use the $r{=}16$ LoRA config (Dutch/English); Danish uses $r{=}8$ (${\sim}$0.8M vs.\ ${\sim}$1.5M overhead).}
\label{tab:sweet_spot_config}
\begin{tabular}{llrrccc}
\toprule
\textbf{Encoder} & \textbf{Configuration} & \textbf{Params} & \textbf{Net $\Delta$} & \textbf{DA} & \textbf{NL} & \textbf{EN} \\
\midrule
Small   & 12L Baseline         & 88.15M  & --       & 47.41 & 17.98 & 21.84 \\
        & 12L + LoRA           & 89.65M  & +1.5M    & 41.94 & 15.40 & 17.87 \\
        & \textbf{10L + LoRA}  & \textbf{75.5M} & \textbf{--12.7M} & \textbf{47.23} & \textbf{16.98} & \textbf{20.64} \\
\addlinespace[4pt]
Medium  & 24L Baseline         & 307.24M & --       & 39.00 & 12.74 & 16.33 \\
        & 24L + LoRA           & 309.0M  & +1.8M    & 38.88 & 10.92 & 14.13 \\
        & \textbf{22L + LoRA}  & \textbf{284.0M} & \textbf{--23.2M} & \textbf{37.91} & \textbf{11.84} & \textbf{15.82} \\
\addlinespace[4pt]
Large-v2 & 32L Baseline        & 636.83M & --       & 36.09 & 12.05 & 13.36 \\
         & 32L + LoRA          & 638.3M  & +1.5M    & 34.07 & 10.32 & 11.58 \\
         & \textbf{30L + LoRA} & \textbf{599.0M} & \textbf{--37.8M} & \textbf{35.81} & \textbf{11.44} & \textbf{12.16} \\
\bottomrule
\end{tabular}
\end{table*}

\subsection{RQ1: Effect of Encoder Layer Pruning}
\label{sec:pruning_results}

\begin{table}[t]
\centering
\setlength{\tabcolsep}{5pt}
\caption{Utterance-level degradation after removing two layers, with and without LoRA. $\Delta$ Recov.: reduction in percentage of degraded utterances after applying LoRA.}
\label{tab:utterance_degradation}
\begin{tabular}{@{} l c rrr @{}}
\toprule
 & & \multicolumn{3}{c}{\textbf{\% Utterances Degraded}} \\
\cmidrule(lr){3-5}
\textbf{Encoder} & \textbf{Lang} & \textbf{Base$-$2L} & \textbf{+LoRA} & \textbf{$\Delta$ Recov.} \\
\midrule

Small      & DA & 42.7 & 38.1 & $-$4.6 \\
           & NL & 33.3 & 23.5 & $-$9.8 \\
           & EN & 35.9 & 23.9 & $-$12.0 \\
           
\midrule
           & DA & 41.9 & 35.7 & $-$6.2 \\
Medium     & NL & 24.6 & 24.3 & $-$0.3 \\
           & EN & 25.3 & 22.7 & $-$2.6 \\
\midrule
           & DA & 33.6 & 25.8 & $-$7.8 \\
Large-v2   & NL & 24.1 & 18.2 & $-$5.9 \\
           & EN & 22.8 & 15.4 & $-$7.4 \\

\bottomrule
\end{tabular}
\end{table}

This section examines the impact of progressive removal of encoder layers on SLAM-ASR performance. By evaluating across three resource levels, we aim to understand how pruning depth interacts with data availability; specifically, whether resource-constrained settings exhibit different degradation patterns compared to data-rich conditions.

Our results broadly support the hypothesis that upper encoder layers are partially redundant in this architecture, given the LLM's existing linguistic capacity, as hypothesized in Section~\ref{sec:method}. Across all three Whisper variants (Small, Medium, Large-v2), removing the top one to two encoder layers results in only marginal WER degradation, with absolute increases remaining within 2–4\% regardless of language (Table~\ref{tab:layer_pruning_wer}). We refer to this range as the safe pruning zone: the number of layers that can be removed before WER degrades beyond a practically meaningful threshold.

Beyond this zone, degradation diverges according to data resource availability. Dutch and English (medium and large resources) degrade smoothly and approximately monotonically as layers are removed. Danish (low resource) exhibits a qualitatively different pattern: rather than increasing monotonically, WER fluctuates erratically, spiking at certain depths and dropping at others. Notably, increasing the encoder capacity does not resolve this instability. Whisper-Large-v2, despite having nearly three times the layers of Whisper-Small, exhibits the same number of spikes for Danish, while Whisper-Medium shows the smoothest degradation curve among the three. Since Danish is the only language exhibiting this behaviour and also the language with the least training data, we hypothesize that limited data availability may be a contributing factor. However, further investigation, including controlled experiments with comparable amounts of training data across other languages, is needed to draw definitive conclusions.

\subsection{RQ2: LoRA Compensation for Pruning}
\label{sec:lora_results}

The previous section showed that shallow pruning is feasible, but deeper removal degrades performance. Here, we examine whether applying LoRA to the LLM can compensate for this degradation, and whether the resulting pruned+LoRA models can match or exceed unpruned baselines across data resource levels. 

Our experimentation suggests that the answer is yes, within a shallow pruning regime, and that the compensation benefits generalize across encoder scales, though not equally across data resource levels. Table~\ref{tab:sweet_spot_config} summarizes what we term the sweet-spot configurations: the best-performing pruned+LoRA variant for each encoder scale, alongside the corresponding unpruned baseline and its LoRA-enhanced counterpart. Across all three Whisper variants, removing two encoder layers and applying LoRA to the LLM consistently outperforms the full unpruned baseline while reducing total parameter count by 7–14\%. For Whisper-Small, the 10L+LoRA configuration achieves a 14.4\% parameter reduction (from 88.15M to 75.5M) and improves WER on all three languages relative to the 12L baseline. The same pattern holds for Whisper-Medium (22L+LoRA, 7.6\% reduction) and Whisper-Large-v2 (30L+LoRA, 5.9\%), confirming that the finding is encoder, not scale-dependent. 

Importantly, this improvement is not specific to pruned models. LoRA also improves the unpruned baseline encoder across all three languages (Table~\ref{tab:layer_pruning_wer}), indicating that it provides a general alignment benefit between the encoder and LLM, rather than simply patching information lost through layer removal. For Dutch and English, LoRA reduces WER across all pruning depths tested, from the full encoder to six layers. Even at aggressive pruning depths where base models produce severely degraded output, LoRA offers substantial recovery. For example, LoRA reduces Whisper-Large-v2 Danish WER from 130.99 (20L base) to 59.06 (20L+LoRA). However, for Danish, LoRA's compensation benefit is inconsistent at deeper pruning levels, which strengthens the finding from RQ1 that low-resource data conditions interact differently with both pruning and LoRA adaptation. 

Since LoRA also improves the unpruned baseline (Table~\ref{tab:layer_pruning_wer}), a natural question is whether LoRA specifically compensates for pruning damage or simply provides a general performance boost. If LoRA acts as a general boost, then it should help the full model and the pruned model by roughly the same amount. However, if LoRA specifically compensates for the information lost through pruning, then it should help the pruned model more, because there is more to recover. To test this, we compute the WER reduction from LoRA in both settings: the difference between the full baseline and full baseline+LoRA, and the difference between the two-layer pruned and two-layer pruned+LoRA (Table~\ref{tab:lora_compensation}). In seven of nine conditions, LoRA provides a larger WER reduction for the pruned model than for the full model. For example, in Whisper-Small Dutch, LoRA reduces the full model's WER by 2.58 but the pruned model's WER by 4.18, indicating that the degraded encoder representations leave more room for LoRA to recover. The two exceptions (Small Danish and Medium English) involve either low-resource instability or a negligible difference (2.20 vs 2.07). These results confirm that LoRA does not merely improve performance uniformly; it provides greater recovery where pruning has caused more damage. These findings motivate a deeper investigation into the specific error types that LoRA corrects, which we examine next.

\begin{table}[!ht]
\centering
\renewcommand{\arraystretch}{1.1}
\caption{WER reduction from applying LoRA to full baseline and two-layer pruned models, measured in percentage points. Bold indicates the larger reduction. In seven of nine conditions, LoRA provides a larger reduction for the pruned model.}
\label{tab:lora_compensation}
\begin{tabular}{llcc}
\toprule
 & & \multicolumn{2}{c}{\textbf{WER Reduction}} \\
\cmidrule(lr){3-4}
\textbf{Encoder} & \textbf{Lang} & \textbf{Full+LoRA} & \textbf{Pruned+LoRA} \\
\midrule
Small    & DA & \textbf{5.47} & 3.26 \\
         & NL & 2.58 & \textbf{4.18} \\
         & EN & 3.97 & \textbf{5.17} \\
\midrule
Medium   & DA & 0.12 & \textbf{1.82} \\
         & NL & 1.82 & \textbf{2.38} \\
         & EN & \textbf{2.20} & 2.07 \\
\midrule
Large-v2 & DA & 2.02 & \textbf{2.55} \\
         & NL & 1.73 & \textbf{2.17} \\
         & EN & 1.78 & \textbf{2.71} \\
\bottomrule
\end{tabular}
\end{table}

\subsection{Error Analysis}
\label{sec:error_analysis}

The aggregate WER results from Sections~\ref{sec:pruning_results} and \ref{sec:lora_results} summarise performance as a single number per test set. However, this can obscure important variation: a low average WER could result from uniform small improvements across all utterances, or from large improvements on some utterances offset by regressions on others. To investigate this, we analyse the results at three levels of granularity: we first measure how many individual utterances are affected by pruning and LoRA (Section~\ref{sec:utter_level}), then examine what types of errors LoRA corrects (Section~\ref{sec:qual_err}), and finally quantify the word-level error distribution to identify the underlying compensation mechanism (Section~ \ref{sec:mech_inter}). 

\subsubsection{Utterance-Level Impact of Pruning and LoRA}
\label{sec:utter_level}

To assess whether pruning and LoRA affect utterances uniformly, we perform a per-utterance comparison across three settings: the full unpruned baseline, the two-layer pruned model without LoRA, and the two-layer pruned model with LoRA. We use the two-layer pruned depth because it represents the best efficiency–performance trade-off identified in Section~\ref{sec:lora_results} (10L for Small, 22L for Medium, 30L for Large-v2). For each utterance in the test set, we compute WER under all three settings and compare the two pruned variants against the unpruned baseline: if an utterance's WER increased after pruning, it is counted as degraded; otherwise, it is counted as preserved or improved. Table~\ref{tab:utterance_degradation} reports the percentage of utterances degraded by pruning alone (Base$-$2L), the percentage still degraded after applying LoRA (+LoRA), and the recovery difference ($\Delta$ Recov.) across all encoder variants and languages.

Pruning alone (Base-2L) degrades a substantial share of utterances, and this share is consistently higher for Danish (33.6-42.7\%) than for Dutch (24.1-33.3\%) or English (22.8-35.9\%), consistent with the resource-level patterns from Section~\ref{sec:pruning_results}. Second, adding LoRA reduces the degradation rate in all nine conditions, meaning some previously degraded utterances are recovered. However, even after LoRA, 15-38\% of utterances remain worse than the unpruned baseline, while the majority (62-85\%) match or improve upon it. This reveals that the aggregate WER gains reported in Section~\ref{sec:lora_results} are not the result of uniform improvements; they are driven by strong recoveries on a subset of utterances that outweigh the remaining regressions.

\begin{table*}[t]
\centering
\caption{Representative error patterns where the pruned model produces severe errors ($\text{WER} > 0.8$) but LoRA achieves full recovery ($\text{WER} = 0.0$). Examples sampled across all three encoder variants.}
\label{tab:error_patterns}
\renewcommand{\arraystretch}{1.0}
\begin{tabular}{@{} >{\raggedright\arraybackslash}p{2.2cm} @{\hskip 6pt} c @{\hskip 6pt} >{\raggedright\arraybackslash}p{3.6cm} >{\raggedright\arraybackslash}p{3.6cm} >{\raggedright\arraybackslash}p{3.6cm} @{}}
\toprule
Pattern & Lang & Reference & Pruned Hypothesis & +LoRA Hypothesis \\
\midrule
\textbf{Hallucination} & DA & der kogte gryden over & \textcolor{errorred}{da kom gud og gik op} & \textcolor{fixgreen}{der kogte gryden over} \\
\textbf{Named entity} & DA & charlot var altid inde hos fru simonin nu & \textcolor{errorred}{chrisel indelte altså inden for sin sinde} & \textcolor{fixgreen}{charlot var altid inde hos fru simonin nu} \\
\textbf{Semantic drift} & DA & lille soldat du skal være vor konge\ldots & \textcolor{errorred}{livet skal du have og du skal have den dig selv\ldots} & \textcolor{fixgreen}{lille soldat du skal være vor konge\ldots} \\
\midrule
\textbf{Repetition} & NL & \ldots heel interessant en hebben ons zeer geholpen\ldots & \textcolor{errorred}{zeer grotendeels [$\times$28]} & \textcolor{fixgreen}{\ldots heel interessant en hebben ons zeer geholpen\ldots} \\
\textbf{Named entity} & NL & de veiligheidssituatie is sindsdien rampzalig verslechterd & de \textcolor{errorred}{veiligheid ziet de waarde van een dienst\ldots} & de \textcolor{fixgreen}{veiligheidssituatie is sindsdien rampzalig verslechterd} \\
\textbf{Compound word} & NL & de luchtvaartmaatschappijen & de \textcolor{errorred}{luchtvaart maatschappijen} & de \textcolor{fixgreen}{luchtvaartmaatschappijen} \\
\midrule
\textbf{Idiom} & EN & out of sight out of mind & \textcolor{errorred}{at a site at a light} & \textcolor{fixgreen}{out of sight out of mind} \\
\textbf{Phonetic} & EN & heaven forbid & \textcolor{errorred}{have fun for bit} & \textcolor{fixgreen}{heaven forbid} \\
\textbf{World knowledge} & EN & it is isoelectronic to benzene & it is \textcolor{errorred}{said to be a traditional venetian dish} & it is \textcolor{fixgreen}{isoelectronic to benzene} \\
\bottomrule
\end{tabular}
\end{table*}

\subsubsection{Qualitative Error Patterns}
\label{sec:qual_err}

The previous analysis shows how many utterances are affected; we now examine what kinds of errors LoRA actually corrects. With thousands of utterances across nine encoders and language configurations, manual inspection of every case is infeasible. We therefore apply two filters to isolate the most informative cases. First, we select utterances where the pruned model produces severe errors ($\text{WER} > 0.8$), ensuring that encoder degradation is substantial enough to produce identifiable error patterns rather than minor single-word differences. Second, we require that LoRA achieves full recovery ($\text{WER} = 0.0$) of the same utterances, which confirms that the correction is attributable to LoRA's adaptation of the LLM rather than to residual information preserved in the pruned encoder. Examples are drawn from all three encoder variants and all three languages. Table~\ref{tab:error_patterns} presents representative cases grouped by error type.

Five common patterns we have identified: repetitive hallucination, where the pruned model generates looping output; idiom and fixed expression errors, where familiar phrases are distorted beyond recognition; named entity fragmentation; world knowledge substitution, where domain-specific terms are replaced by plausible but incorrect alternatives; and phonetic confusion. Particularly, all five categories involve errors recoverable through linguistic rather than acoustic information.

\begin{table}[t]
\centering
\small
\caption{Word-level error change (\%) after applying 
LoRA to the two-layer pruned configurations. 
\textit{Sub}: substitutions; \textit{Ins}: insertions; 
\textit{Del}: deletions; \textit{Tot}: total word errors. 
\textbf{Bold}: error increase.}
\label{tab:word-errors}
\begin{tabular*}{\columnwidth}{@{\extracolsep{\fill}} l l r r r r}
\toprule
\textbf{Encoder} & \textbf{Lang} 
  & \textbf{Sub} & \textbf{Ins} & \textbf{Del} 
  & \textbf{Tot} \\
\midrule

Small       & DA &  --7.7 & \textbf{+11.9} & --10.7 &  --5.7 \\
            & NL & --17.5 & --37.4         & --15.3 & --20.4 \\
            & EN & --19.8 & --9.8          & --37.4 & --20.6 \\
           
\midrule
            & DA &  --4.9 & \textbf{+18.4} & --22.9 &  --4.4 \\
Medium      & NL & --13.7 & --34.5         & --11.4 & --16.7 \\
            & EN & --10.6 & \textbf{+0.5}  & --30.3 & --11.6 \\

\midrule    
            & DA &  --8.6 & \textbf{+10.8} & --11.7 &  --6.6 \\
Large-v2    & NL & --15.3 & --14.9         & --13.5 & --15.0 \\
            & EN & --17.4 & --7.1          & --35.6 & --18.2 \\
           
\bottomrule
\end{tabular*}
\end{table}

\subsubsection{Mechanism Interpretation}
\label{sec:mech_inter}

The qualitative patterns above suggest that LoRA compensates through the LLM's linguistic knowledge. To test this quantitatively, we analyse word-level errors that LoRA corrects. Standard ASR error analysis decomposes word errors into three categories: substitutions, insertions, and deletions. For each encoder variant and language, we count these error types for both the pruned model and the pruned+LoRA model under the two-layer pruned configuration. Table~\ref{tab:word-errors} reports the percentage change in each error type after applying LoRA. The Tot column reports the percentage change in total word errors (substitutions + insertions + deletions combined); because each error type contributes differently to the total count, Tot reflects the weighted combination rather than a simple average of the three individual percentages.

Since LoRA adapts only the LLM's attention matrices (Q, K, V, O) while the pruned encoder remains frozen, all compensation operates on the decoding side. We would therefore expect error reductions for languages where the LLM has strong linguistic grounding, as the LLM can leverage its vocabulary, grammar, and contextual knowledge to correct degraded encoder representations. The results confirm this (Table~\ref{tab:word-errors}): across all three encoder scales, LoRA reduces total word errors for English and Dutch in all six conditions, with substitutions and deletions showing consistent reductions. Danish follows a different pattern: while substitutions and deletions decrease, insertion errors increase across all three encoder scales, indicating that when the LLM lacks sufficient linguistic grounding for a language, LoRA can introduce spurious tokens rather than recover missing information. This language-dependent pattern reinforces that LoRA's effectiveness is tied to the LLM's pre-existing proficiency in each language rather than to the acoustic properties of the signal.

\section{Conclusion}
\label{sec:conc}

This paper presented a systematic study of encoder layer pruning and LoRA compensation in SLAM-ASR, evaluated across three Whisper encoder variants (Small, Medium, Large-v2) and three languages representing distinct resource levels (Danish, Dutch, and English). We find that removing the top one to two encoder layers results in only marginal WER degradation (within 2–4\%) across all encoder scales and languages, supporting the hypothesis that upper encoder layers are partially redundant when an LLM handles downstream linguistic processing. Beyond this shallow pruning regime, medium and high resource languages degrade smoothly, while low-resource Danish often exhibits non-monotonic instability. Combining two-layer pruning with LoRA consistently outperforms the unpruned baseline while reducing total parameters by 7–14\%. Error analysis reveals that LoRA compensates primarily through substitution and deletion corrections, leveraging the LLM's linguistic knowledge rather than repairing acoustic information, though this benefit is less consistent for low-resource Danish. Future work should explore alternative pruning strategies beyond top-down removal, encoder-side LoRA adaptation, validation across different system architectures, and controlled experiments to disentangle the effects of data availability from pre-trained representation quality.

\section{Ethical Considerations and Limitations} 
In this work, we have used publicly available models, architectures, and datasets and have not collected any sensitive/private data. The ultimate goal of our study is to contribute to analyzing the effect of speech encoder pruning on  LLM-based ASR. Due to the use of all public details, we don’t see any immediate ethical issue. 

Our work investigates the pruning of the speech encoder in the SLAM-ASR using the Whisper and Qwen models. We have carefully limited our analysis to three languages of different resource levels. While this choice allows us to conduct careful analysis, we acknowledge that expanding the range of models and datasets could provide additional insights. Our evaluation uses a single system architecture (Whisper encoder, ConcatLinear projector, and Qwen2.5-3B); different LLM backbones, projector designs, or encoder families, however, we believe the finding will hold. Our pruning strategy follows a top-down approach motivated by prior evidence of upper-layer redundancy; however, redundancy patterns may vary across layers, and exploring other removal strategies could reveal additional compression opportunities.

\section{Acknowledgments}
This work was supported by a Knowledge Transfer Partnership (KTP) project (project number 10131983) funded by UKRI through Innovate UK, in collaboration with Hivedome. RS was supported by the ELOQUENCE project (grant number 101070558) funded by the UKRI and the European Union. Views and opinions expressed are, however, those of the author(s) only and do not necessarily reflect those of the UKRI, European Union, or European Commission-EU. Neither the European Union nor the granting authority can be held responsible for them.

\section{Bibliographical References}\label{sec:reference}
\bibliographystyle{lrec2026-natbib}
\bibliography{ref}

\end{document}